\documentclass[runningheads]{llncs}

\PassOptionsToPackage{table,xcdraw}{xcolor}
\usepackage{eccv}
\usepackage{array}
\usepackage{tabularx}
\usepackage{booktabs}
\usepackage{threeparttable}

\usepackage{eccvabbrv}
\usepackage{tcolorbox}
\usepackage{graphicx}
\usepackage{booktabs}

\usepackage[accsupp]{axessibility}  

\usepackage[breaklinks,colorlinks,citecolor=eccvblue]{hyperref}

\usepackage{orcidlink}

\begin{document}

\title{Team PA-VCG's Solution for Competition on Understanding Chinese College Entrance Exam Papers in ICDAR'25} 

\titlerunning{Team PA-VCG's Solution}

\author{Wei Wu\inst{1}$^{\dagger}$ \and
Wenjie Wang\inst{1}$^{\dagger}$ \and
Yang Tan\inst{1}$^{\dagger}$ \and
Ying Liu\inst{1} \and
Liang Diao\inst{1}$^{*}$  \and Lin Huang\inst{1} \and Kaihe Xu\inst{1} \and Wenfeng Xie\inst{1} \and Ziling Lin\inst{1}  \\%
\hspace{1em}
\email{diaoliang91@gmail.com}}

\authorrunning{Wei Wu, Wenjie Wang et al.}


\institute{Visual Computing Group, Ping An Property \& Casualty Insurance Company}

\maketitle
\footnotetext{$^{\dagger}$Main Contribution \quad $^{*}$Corresponding author}

\begin{abstract}
This report presents Team PA-VGG's solution for the ICDAR'25 Competition on Understanding Chinese College Entrance Exam Papers. In addition to leveraging high-resolution image processing and a multi-image end-to-end input strategy to address the challenges of dense OCR extraction and complex document layouts in Gaokao papers, our approach introduces domain-specific post-training strategies. Experimental results demonstrate that our post-training approach achieves the most outstanding performance, securing \textbf{first place with an accuracy rate of 89.6\%.}
\end{abstract}

\begin{center}
{\footnotesize
\textbf{Code:} \textcolor{gray}{\texttt{\url{https://github.com/Dejavuvvw/1st_solution_UCCEEP_ICDAR2025}}} \\
\textbf{Model:} \textcolor{gray}{\texttt{\url{https://huggingface.co/Dejavuvvw/1st_solution_UCCEEP_ICDAR2025}}}
}
\end{center}

\section{Competition Setting}

This competition tests MLLMs' skill in analyzing Chinese Gaokao exam papers. To do well, models need to handle both detailed OCR processing for accurate text extraction and complex document layouts, like charts, tables, and multi-column formats, all at the same time.

\section{Method}

\subsection{High resolution for dense OCR prediction} 

The competition entails analyzing dense Chinese Gaokao exam paper images, each containing about \textbf{1,000 text tokens}. Recent studies \cite{bai2025qwen2,chen2024internvl,yao2024minicpm} show that such token-dense scenarios require \textbf{high-resolution visual processing} to tackle OCR-intensive tasks. Prior work confirms that effective visual encoding boosts performance, especially in retaining fine-grained text and structure.

\begin{table}[htb!]
  \caption{Training hyper-parameters}
  \label{tabel1}
  \centering
  \begin{tabular}{@{}lll@{}}
    \toprule
    Hyper-paramerter & Lora & Full \\
    \midrule
    batch-size  & 16 & 64 \\
    learning rate & 5e-5 & 6e-6 \\
    learning rate schedule & cosine & cosine \\
    learning rate warm-up ratio & 0 & 0 \\
    weight decay & 0.1 & 0.1 \\
    grad norm clipping & 1 & 1 \\
    epoch & 10 & 5 \\
    optimizer & AdamW & AdamW \\
    float precision & bfloat16 & bfloat16 \\
    deepspeed configuration & - & zero3-offload \\
  \bottomrule
  \end{tabular}
\end{table}

\subsection{Multiple images MLLM} 
Unlike methods that process a single image at a time followed by reasoning, our approach directly inputs all task-related images and prompts end-to-end into MLLM. This allows the model to fully leverage image feature information across multiple images, enhancing its layout, text, and image feature extraction capabilities.
\subsection{MLLM Finetuning}
\subsubsection{Distillation from OCR foundation MLLM}
To avoid a decline in the generalization ability of the multimodal large model and ensure the basic OCR capability is preserved during fine-tuning for downstream tasks, we utilize OVIS2 \cite{lu2024ovis}, a foundation OCR MLLM, to extract text information from the data. Subsequently, this text information is combined with task-specific data as distillation data for post-training.
\subsubsection{Lora Finetune}
Lora \cite{hu2022lora}, a commonly used method for specific tasks, efficiently adapts pre-trained models by adding a few trainable parameters. It is known for its fast convergence and ability to prevent overfitting. We integrated Lora adapters into key layers of the models. Experimental results demonstrate that Lora fine-tuning enhances the models' ability to answer the competition question.
\subsubsection{Full-parameter Finetune}
To further explore the potential accuracy limits, we conducted full-parameter fine-tuning on the models. This approach, unlike Lora, involves updating all model parameters, which consequently demands more computational resources. Conventionally, fine-tuning a large model on a limited dataset is prone to overfitting. Surprisingly, our experiments showed that full-parameter fine-tuning avoided overfitting and significantly outperformed the Lora method. The underlying reasons for this unexpected outcome require further investigation.
\subsection{Prompt Refinement and Post Processing}

\subsubsection{Page-Aware Prompt Refinement}
By analysis of the error samples of the baseline model, we find that the MLLM still has unclear recognition of page numbers. Therefore, we explicitly add page number indications in the prompt. The prompt is as follows: 
\begin{tcolorbox}  
Page1:<image>,Page2:<image>,Page3:<image>. \textcolor{blue}{\{question\}}.
\end{tcolorbox}
Besides, for different questions, we refine the system prompts to supply necessary prior knowledge and rule constraints for answering.
\subsubsection{Process Reasoning Refinement}
We observe that with direct end-to-end training, the model fails to implicitly learn the intermediate reasoning process. For example, we found that MLLMs perform poorly in determining whether the answer to a given question is contained within a single page and in predicting the page number where the answer first appears. To address these limitations, we introduce a \textbf{process reasoning refinement} by generating intermediate-step Q\&A augmented data, such as asking ``Is the answer to question~12 split across multiple pages?'' and ``Please provide, in Arabic numerals, the page number where the answer section first appears.''

\subsubsection{Post process}
To regulate the output of MLLM, several post-processing steps are implemented. For instance, we eliminate blank characters and any text unrelated to the answers.

\section{Implementation Details and Experiment}
\begin{table}[htb]
  \centering
  \caption{Compatible Table with Row Coloring}
  \label{tab:ablation}
  \renewcommand{\arraystretch}{1.2}
  \begin{tabular}{lccc}
    \toprule
    \textbf{Model} & \textbf{Config} & \textbf{Params} & \textbf{Acc (\%)} \\
    \midrule
    QwenVL-2.5 (3B) & LoRA baseline & 3B & 75.0 \\
    QwenVL-2.5 (7B) & LoRA baseline & 7B & 75.6 \\
    QwenVL-2.5 (3B) & + Prompt Refine + Distil & 3B & 76.6 \\
    \midrule
    Ovis-2 (4B) & Full fine-tuning & 4B & 70.0 \\
    QwenVL-2.5 (7B) & Full (lr=1e-6) & 7B & 82.2 \\
    QwenVL-2.5 (7B) & Full (lr=6e-6) & 7B & 88.3 \\
    \rowcolor[HTML]{F2F2F2}
    QwenVL-2.5 (7B) & \textbf{Full + PR + Distil} & 7B & \textbf{89.6} \\
    \bottomrule
  \end{tabular}
\end{table}

Considering the GPU memory constraints, we excluded training data with more than 25 images. Our training code leverages the widely-used Transformers \cite{wolf2020transformers} and ms-swift \cite{zhao2024swiftascalablelightweightinfrastructure} libraries. We conducted experiments on various base models, including Qwen2.5 VL 3B, Qwen2.5 VL 7B, and Ovis2 4B. For Lora fine-tuning, we used a single A800-80G GPU, whereas for full-parameter fine-tuning, we employed eight A800-80G GPUs. 

The detailed training hyper-parameters are shown in Table~\ref{tabel1}. We conducted extensive ablation studies to validate our design choices. As shown in Table~\ref{table:ablation}, our full-parameter fine-tuning approach with learning rate 6e-6 achieves the best performance when combined with prompt refinement and OCR distillation. The final competition results validate our methodology, as shown in Table~\ref{table}.




\begin{table}[htb!]
  \caption{Final Competition Team Rankings}
  \label{table}  
  \centering
  \setlength{\tabcolsep}{12pt}  
  \renewcommand{\arraystretch}{1.2}  
  \begin{tabularx}{0.8\linewidth}{@{} l >{\raggedright}X c @{}}  
    \toprule[1.5pt]  
    \textbf{Rank} & \textbf{Team Name} & \textbf{Acc. (\%)} \\  
    \midrule[1pt]  
     1 & PA-VGG         & 89.6 \\
     2 & PAL Team       & 82.1 \\
     3 & 360ailab       & 78.8 \\
     4 & PKU-CNKI       & 75.9 \\
     5 & GlintAnything  & 74.0 \\
     6 & TeamCufe       & 51.5 \\
    \bottomrule[1.5pt]  
  \end{tabularx}
\end{table}

\par\vfill\par

\clearpage  

%
%
\bibliographystyle{splncs04}
\bibliography{main}

\begin{thebibliography}{1}
\providecommand{\url}[1]{\texttt{#1}}
\providecommand{\urlprefix}{URL }
\providecommand{\doi}[1]{https://doi.org/#1}

\bibitem{bai2025qwen2}
Bai, S., Chen, K., Liu, X., Wang, J., Ge, W., Song, S., Dang, K., Wang, P., Wang, S., Tang, J., et~al.: Qwen2. 5-vl technical report. arXiv preprint arXiv:2502.13923  (2025)

\bibitem{chen2024internvl}
Chen, Z., Wu, J., Wang, W., Su, W., Chen, G., Xing, S., Zhong, M., Zhang, Q., Zhu, X., Lu, L., et~al.: Internvl: Scaling up vision foundation models and aligning for generic visual-linguistic tasks. In: Proceedings of the IEEE/CVF conference on computer vision and pattern recognition. pp. 24185--24198 (2024)

\bibitem{hu2022lora}
Hu, E.J., Shen, Y., Wallis, P., Allen-Zhu, Z., Li, Y., Wang, S., Wang, L., Chen, W., et~al.: Lora: Low-rank adaptation of large language models. ICLR  \textbf{1}(2), ~3 (2022)

\bibitem{lu2024ovis}
Lu, S., Li, Y., Chen, Q.G., Xu, Z., Luo, W., Zhang, K., Ye, H.J.: Ovis: Structural embedding alignment for multimodal large language model. arXiv preprint arXiv:2405.20797  (2024)

\bibitem{wolf2020transformers}
Wolf, T., Debut, L., Sanh, V., Chaumond, J., Delangue, C., Moi, A., Cistac, P., Rault, T., Louf, R., Funtowicz, M., et~al.: Transformers: State-of-the-art natural language processing. In: Proceedings of the 2020 conference on empirical methods in natural language processing: system demonstrations. pp. 38--45 (2020)

\bibitem{yao2024minicpm}
Yao, Y., Yu, T., Zhang, A., Wang, C., Cui, J., Zhu, H., Cai, T., Li, H., Zhao, W., He, Z., et~al.: Minicpm-v: A gpt-4v level mllm on your phone. arXiv preprint arXiv:2408.01800  (2024)

\bibitem{zhao2024swiftascalablelightweightinfrastructure}
Zhao, Y., Huang, J., Hu, J., Wang, X., Mao, Y., Zhang, D., Jiang, Z., Wu, Z., Ai, B., Wang, A., Zhou, W., Chen, Y.: Swift:a scalable lightweight infrastructure for fine-tuning (2024), \url{https://arxiv.org/abs/2408.05517}

\end{thebibliography}
\end{document}